\documentclass[10pt,twocolumn,letterpaper]{article}

\usepackage{cvpr}              
\definecolor{cvprblue}{rgb}{0.21,0.49,0.74}
\usepackage[pagebackref,breaklinks,colorlinks,allcolors=cvprblue]{hyperref}
\usepackage{times}  
\usepackage{graphicx} 
\usepackage{natbib}  
\usepackage{caption} 
\usepackage{algorithm}
\usepackage{algorithmic}

\usepackage{enumitem}
\usepackage{amssymb}
\usepackage{pifont}
\usepackage{xspace}
\usepackage{amsmath}
\usepackage[dvipsnames]{xcolor}
\usepackage{booktabs}       
\usepackage{colortbl} 
\usepackage{multirow}
\usepackage{newfloat}
\usepackage{listings}
\usepackage{placeins}
\usepackage{stfloats}
\usepackage{float} 
\newcommand{\name}{SRA 2~}
\newcommand{\namex}{SRA 2}
\def\pz{{\phantom{0}}}
\usepackage{tikz}

\usepackage[table]{xcolor}         
\definecolor{tabhighlight}{HTML}{e5e5e5}

\def\paperID{} 
\def\confName{CVPR}
\def\confYear{2026}

\title{SRA 2: Variational Autoencoder Self-Representation Alignment \\ for Efficient Diffusion Training}
\author{Mengmeng Wang\textsuperscript{1} \hspace{5pt} Dengyang Jiang\textsuperscript{2} \hspace{5pt} Liuzhuozheng Li\textsuperscript{2} \hspace{5pt} Yucheng Lin\textsuperscript{1} \hspace{5pt} Guojiang Shen\textsuperscript{1}\thanks{Corresponding authors}\\Xiangjie Kong\textsuperscript{1} \hspace{5pt} Yong Liu\textsuperscript{3} \hspace{5pt} 
Guang Dai\textsuperscript{2} \hspace{5pt} Jingdong Wang\textsuperscript{4}\footnotemark[1] \hspace{5pt} \\
\textsuperscript{1}Zhejiang University of Technology \quad
$^2$SGIT AI Lab, State Grid Corporation of China\\
$^3$ Zhejiang University
$^4$Baidu \quad\\
\vspace{5pt}}

\begin{document}
\twocolumn[{%
\renewcommand\twocolumn[1][]{#1}%
\maketitle

\vspace{-10mm}
\begin{center}
    \centering
    \captionsetup{type=figure}
    \includegraphics[width=1\linewidth]{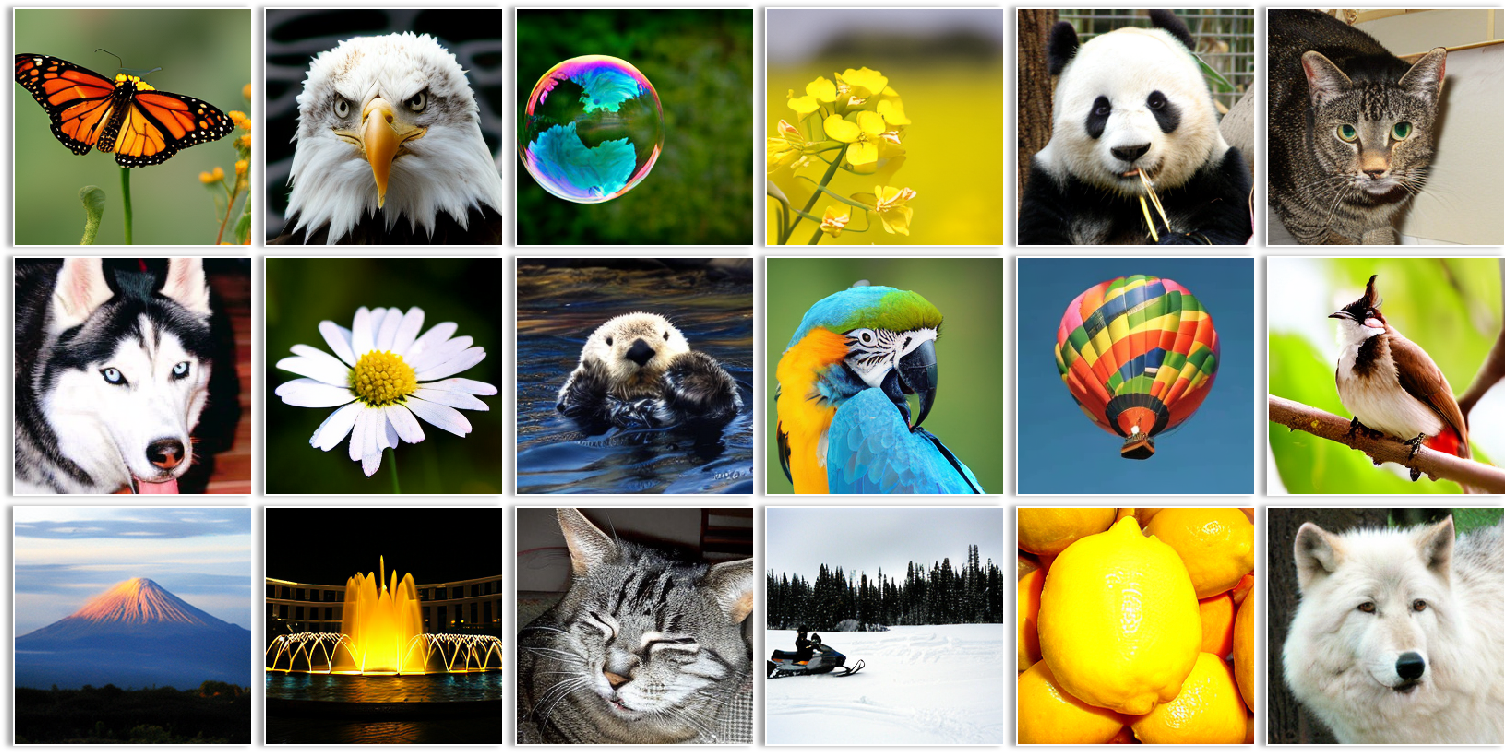} 
  \caption{\textbf{High-Quality Image Generation via \namex}: Samples from SiT-XL/2+\name on ImageNet 256×256. Our method generates images with high fidelity, fine-grained detail, and strong semantic coherence, demonstrating excellent generation quality.}\label{fig:samples}
\end{center}%
}]

\begingroup
\renewcommand\thefootnote{*}
\footnotetext[1]{Corresponding authors.}
\endgroup

\maketitle
\begin{abstract}
Denoising-based diffusion transformers, despite their strong generation performance, suffer from inefficient training convergence. Existing methods addressing this issue, such as REPA (relying on external representation encoders) or SRA (requiring dual-model setups), inevitably incur heavy computational overhead during training due to external dependencies. To tackle these challenges, this paper proposes \textbf{\namex}, a lightweight intrinsic guidance framework for efficient diffusion training. \name leverages off-the-shelf pre-trained Variational Autoencoder (VAE) features: their reconstruction property ensures inherent encoding of visual priors like rich texture details, structural patterns, and basic semantic information. Specifically, \name aligns the intermediate latent features of diffusion transformers with VAE features via a lightweight projection layer, supervised by a feature alignment loss. This design accelerates training without extra representation encoders or dual-model maintenance, resulting in a simple yet effective pipeline. Extensive experiments demonstrate that \name improves both generation quality and training convergence speed compared to vanilla diffusion transformers, matches or outperforms state-of-the-art acceleration methods, and incurs merely 4\% extra GFLOPs with zero additional cost for external guidance models.
\end{abstract}    
\section{Introduction}
\label{sec:intro}
\begin{figure*}
  \centering
  \includegraphics[width=1\linewidth]{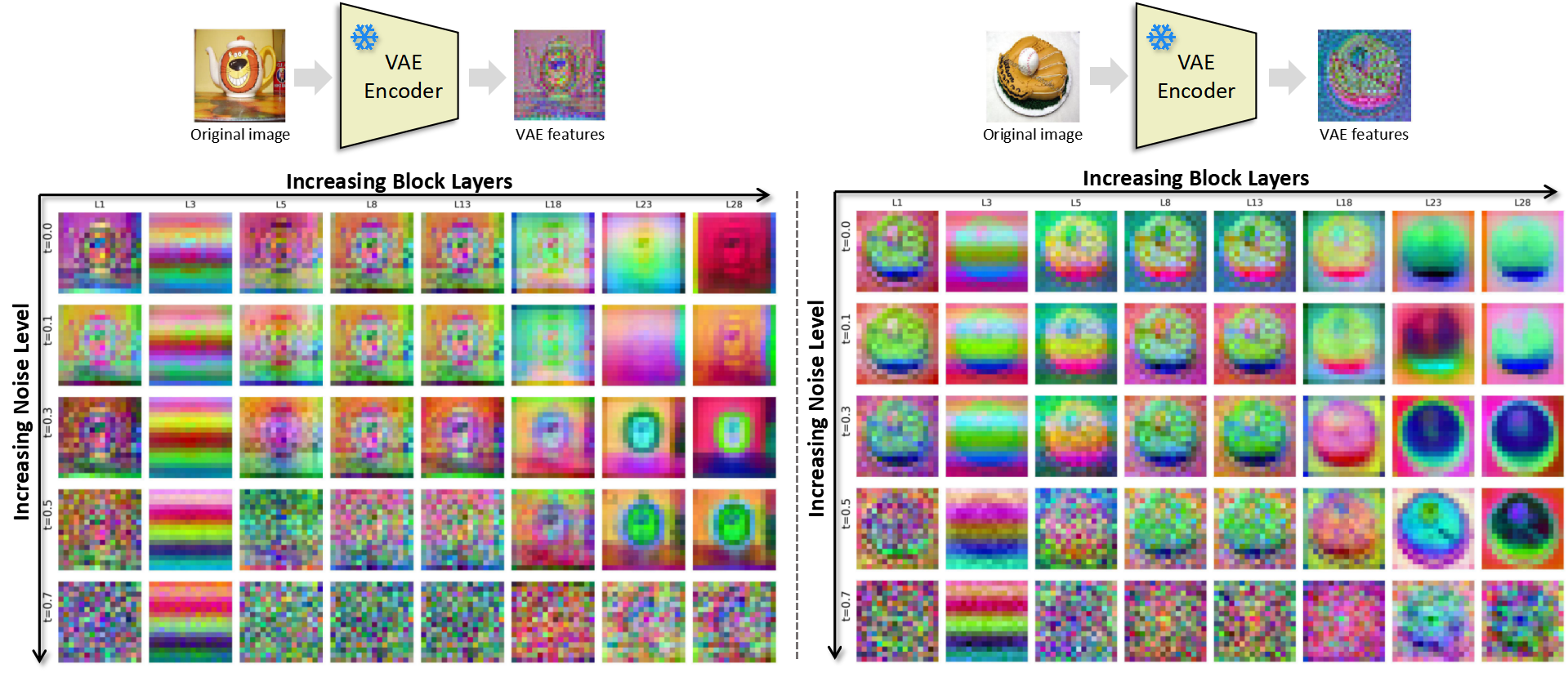} 
  \caption{We empirically visualize the feature information richness of SD-VAE~\cite{sd-vae} and SiT-XL/2~\cite{sit} via PCA~\cite{pca}. \textbf{Top}: VAE features, extracted from original images by an SD-VAE encoder. \textbf{Bottom}: Latent features of SiT-XL/2 across different block layers and noise levels. We observe that SD-VAE features are significantly superior in delineating visual concepts compared to SiT’s latent representations, maintaining clearer details, structural integrity, and stronger semantic coherence. This motivates our use of VAE features for representation alignment.}\label{fig:pca}
  \vspace{-.2cm}
\end{figure*}
Denoising-based generative models, particularly diffusion transformers~\cite{dit,sit}, have attracted significant attention for their exceptional ability to generate diverse, high-fidelity images. They have proven to deliver strong performance across a wide range of domains, including text-to-image synthesis~\cite{pixartalpha,flux}, text-to-video generation~\cite{wan,cogvideox}, image editing~\cite{insertanything,dit4edit}, and 3D asset generation~\cite{direct3d,hunyuan3d}, and so on. However, even the most popular Latent Diffusion Model (LDM)~\cite{ldm} architecture still grapples with the critical issue of slow training convergence, often requiring an enormous number of iterations to achieve satisfactory performance.

To explore the efficient training of large transformer-based diffusion models, self-supervised learning techniques such as masked modeling~\cite{mae} have been adopted~\cite{maskdit,sddit}, which help reduce training costs and accelerate convergence to a certain extent, but they come at the cost of network architecture adjustments such as the need for an additional diffusion decoder. Recently, several methods~\cite{ddt,repa,videorepa,reg} represented by REPA~\cite{repa} have started guiding the optimization of diffusion models by incorporating external large-scale pre-trained representative encoders like DINOv2~\cite{dinov2}. However, despite their promising performance, we observe that integrating these extra representation encoders not only increases training computational overhead but also creates dependencies on external large-scale pre-trained models. In practice, suitable pre-trained models are not available across all domains, such as the video domain or certain specialized downstream tasks, where there is a lack of encoders with strong generalization capabilities. This will significantly limit the applicability of such methods. Therefore, attempts have also been made to accelerate training by leveraging the diffusion transformer’s inherent discriminative information~\cite{sra,ssd}, such as that from different layers or time steps. Nevertheless, their additional cost lies in the need to maintain an extra teacher diffusion model to provide self-alignment guidance during training. Thus, we wonder: \textbf{\textit{Does there exist a simpler and more lightweight guidance approach that can avoid external representative encoders or dual-model maintenance?}}
\begin{figure*}
  \centering
  \includegraphics[width=1.0\linewidth]{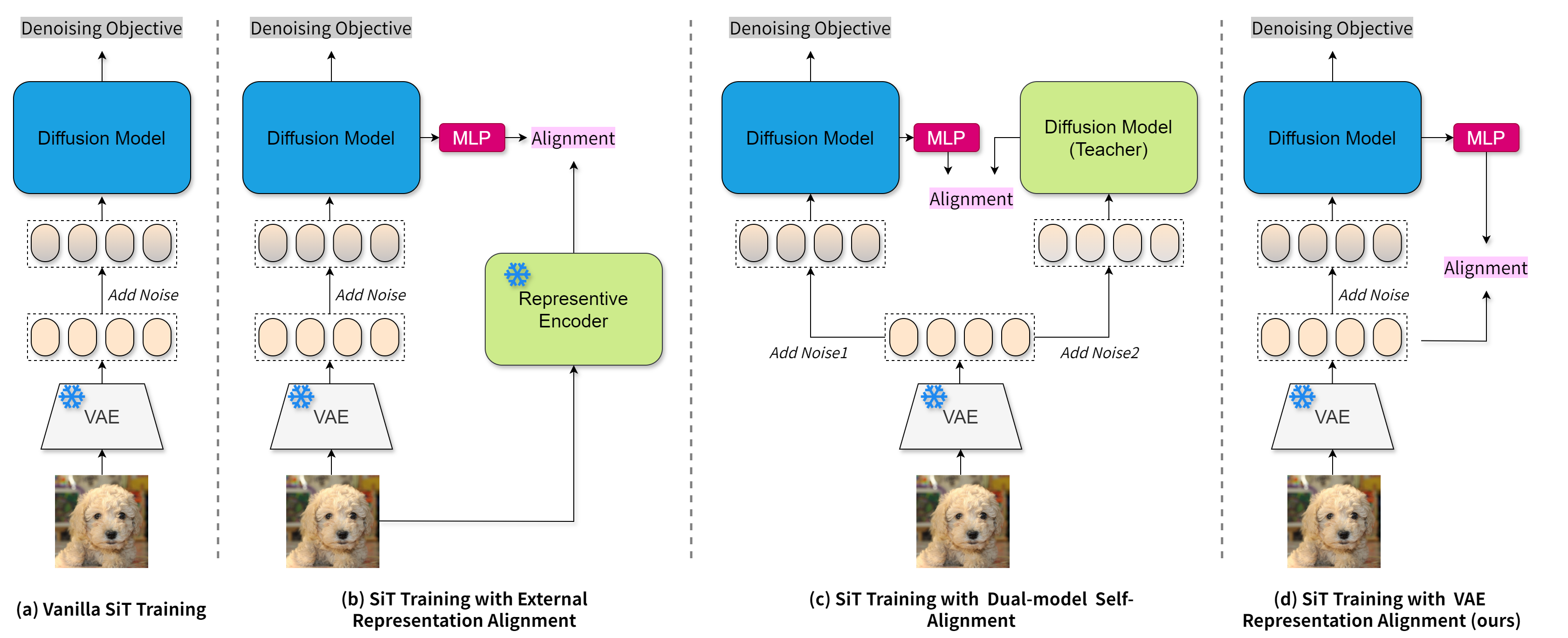}
    \caption{\textbf{Comparison of typical SiT training paradigms.} \textbf{(a)} Vanilla SiT Training: Images are encoded by a VAE, added with noise, and processed by the diffusion model for denoising. \textbf{(b)} SiT Training with External Representation Alignment (e.g., REPA~\cite{repa}): SiT training augmented with an external representative encoder and an MLP for alignment. \textbf{(c)} SiT Training with Dual-model Self-Alignment (e.g., SRA~\cite{sra}): SiT training leveraging a dual-model setup with an MLP for self-alignment, guided by a teacher diffusion model. \textbf{(d)} SiT Training with VAE Representation Alignment (ours): SiT utilizes VAE features as representation guidance and an MLP for alignment, efficiently combining VAE’s semantic richness with SiT’s denoising capability without introducing additional heavy models.} 
    \label{fig:paradigms}
  \vspace{-.2cm}
\end{figure*}

In this paper, we focus our solution on finding more suitable supervisory features, and in this process, we turn to a readily available resource: pre-trained Variational Autoencoders (VAEs). This may be a promising solution to addressing the aforementioned pain point. In the two-stage LDM, the pre-trained VAE in the first stage, having been trained on large-scale natural image datasets, possesses inherent feature encoding capabilities. It can be used for high-quality image reconstruction, thus ensuring that its encoded features encapsulate the image’s texture details, low-level structural patterns, and basic semantic information, as illustrated in Fig.~\ref{fig:pca}. Importantly, these VAE features are usually pre-extracted offline for training the diffusion models in the second stage; thus, they can be conveniently reused directly as our target built-in guidance at no extra cost.
If we can leverage the prior information in VAE features to provide diffusion models with feature-rich, noise-free targets, we can thereby fundamentally help the model improve both optimization efficiency and generation quality.

To address this, we propose \textbf{\namex}, a lightweight intrinsic guidance framework \textbf{that aligns VAE features for self-representation} during the training of diffusion transformer models. Specifically, it leverages off-the-shelf pre-trained VAE features to guide the intermediate layer representations of diffusion models. The intermediate diffusion latent feature is first passed through a projection layer to perform the nonlinear and dimension transformation of the feature space, and then aligned with the target VAE features. To achieve effective alignment, \name incorporates a simple yet efficient feature alignment loss, which minimizes the discrepancy between the diffusion model’s intermediate features and the VAE’s representations. With this simple design, \name keeps the overall training framework highly concise and lightweight as shown in Fig.~\ref{fig:paradigms}(d), eliminating the need for extra representative encoders (as in Fig.~\ref{fig:paradigms}(b)) or dual-model maintenance (as in Fig.~\ref{fig:paradigms}(c)) by directly reusing pre-extracted VAE features. Finally, we conduct extensive experiments to demonstrate the effectiveness of our \name by applying it to the recent diffusion transformer SiT~\cite{sit}. It achieves notable improvements in both generation quality and training convergence speed, while incurring zero additional guidance feature extraction cost and with only 4\% extra GFLOPs for feature alignment during the training process.

The main contributions can be summarized as:
\begin{itemize}
    \item We discover that the features of the pre-trained VAE, by virtue of their reconstruction property, inherently encode rich visual priors, which can serve as a readily available guidance source for diffusion transformer training.
    \item We propose \textbf{\namex}, a simple and lightweight built-in guidance framework that leverages off-the-shelf pre-trained VAE features to align diffusion transformer's intermediate representations, avoiding external model dependencies.
    \item On the ImageNet 256×256 benchmark, our proposed \name achieves notable improvements over the vanilla SiT and matches or surpasses the performance of methods with external model dependencies, while incurring zero additional cost for guidance feature extraction.
\end{itemize}

\section{Related Works}
\label{sec:related}
\noindent\textbf{Diffusion Transformers for image generation.}
Diffusion models have emerged as a powerful paradigm for high-fidelity image generation, evolving from early pixel-space approaches~\cite{adm,vdm,simplediff,cdm} to latent diffusion frameworks (LDM)~\cite{ldm}. With the integration of transformer architectures~\cite{attention,vit}, diffusion transformers~\cite{sd3,lumina2,pixartalpha,flux,dit,sit} have further advanced this paradigm: they leverage transformer-based attention mechanisms to model complex semantic dependencies, enabling scalable high-resolution generation and more sophisticated task adaptation. Foundational works like DiT~\cite{dit} first demonstrated that transformers can model diffusion’s denoising dynamics in latent spaces with minimal structural overhead, while SiT~\cite{sit} advanced this paradigm via linear flow diffusion to balance scalability and performance. Yet a critical limitation remains: their strong performance relies on massive training iterations, slowing convergence. Our work addresses this inefficiency with \namex, a compact method that accelerates training while preserving modern diffusion transformers' architectural elegance.

\noindent\textbf{Diffusion Training Guided by External Dependency.}
Diffusion training guided by external dependencies has provided an effective pathway for improving convergence efficiency and generative quality, with three distinct paradigms: masking modeling~\cite{mdt,mdtv2,maskdit,sddit}, pre-trained representation guidance~\cite{rcg,ddt,repa,reg}, and self-alignment~\cite{dod,sra,ssd}. Masking modeling methods exemplified by methods like SD-DiT~\cite{sddit} and MaskDiT~\cite{maskdit} introduce structured noise through partial masking of input tokens or latent features, creating an external diffusion decoder dependency on reconstructing missing information. Pre-trained representation guidance methods leverage extra features from external foundation models~\cite{dinov2,moco} as a dependency, explicitly aligning the diffusion model’s intermediate representations with these robust, pre-learned representation priors. For instance, RCG~\cite{rcg} uses a pretrained self-supervised encoder~\cite{moco} to map image distributions to aligned representation distributions, while REPA~\cite{repa} and its successors like REG~\cite{reg} and DDT~\cite{ddt} improve semantic representation quality via feature alignment between early diffusion layers and pretrained vision features of DINOv2~\cite{dinov2}. Self-alignment approaches like SRA~\cite{sra} and SSD~\cite{ssd} employ a dual-model framework with an online teacher model that supplies higher quality features characterized by more robust semantics or lower noise. While these methods do accelerate training, they often involve external dependencies that are not always accessible and incur additional training computational overhead. In contrast, our work eschews such external dependencies, aiming for a lightweight intrinsic solution that preserves the simplicity of vanilla diffusion frameworks while achieving comparable or superior acceleration.

\noindent\textbf{Leveraging VAEs for Diffusion Acceleration.} Stable Diffusion~\cite{sd-vae} popularized the latent diffusion paradigm, employing a VAE~\cite{vae}, termed SD-VAE, to encode images into a compact latent space and decode latent tokens back into images. This design enables efficient training and scaling of diffusion models, making it a dominant choice for visual generation. We focus on related methods that optimize VAEs for accelerating diffusion models, which can be categorized into two types. The first type is latent compression methods~\cite{dc-ae,dc-ae2}, which emphasize deeper compression and structured latent subspaces to reduce computation. The second type involves representation alignment for VAE refinement~\cite{vavae,xu2025exploring,repa-e}, which aligns VAE features with pre-trained representations to speed up convergence via semantic priors. For example, REPA-E~\cite{repa-e} attempts to train latent diffusion models together with the VAE tokenizer in an end-to-end manner using representation alignment loss. VAVAE~\cite{vavae} mitigates the optimization dilemma between VAE reconstruction and diffusion generation by aligning VAE representations with external semantic priors. In contrast, our work differs by adopting a sample alignment strategy compatible with SD-VAE, directly leveraging pre-extracted VAE features to enhance the diffusion model's convergence speed without extra VAE training or compression.

\section{Method}
\label{sec:related}
\subsection{Preliminaries}
We apply our \name upon Scalable Interpolant Transformers (SiT)~\cite{sit}, a unified framework that bridges flow and diffusion models using interpolating processes. We focus on its core mechanisms relevant to our method, and briefly elaborate on its foundational principles for clarity.

Here we denote $\boldsymbol{y}_t \in \mathbb{R}^n$ as a sample at time $t$, and SiT models the probability density flowing from a data sample $\boldsymbol{z}\sim p_{data}(\boldsymbol{z})$ to Gaussian noise $\boldsymbol{\epsilon} \sim \mathcal{N}(\boldsymbol{\epsilon} | \boldsymbol{0}, \boldsymbol{I})$ through a time-dependent interpolation:
\begin{equation}
\boldsymbol{y}_t = a_t \boldsymbol{z} + b_t \boldsymbol{\epsilon},
\end{equation}
where $t \in [0, T]$ is the time variable, $a_t$ decreases from 1 to 0, and $b_t$ increases from 0 to 1 (generalizing both flow models, which use finite intervals, and diffusion models, which extend to $T \to \infty$).

The key insight is that sampling can be driven by a \textit{velocity function} $\boldsymbol{v}(\boldsymbol{y}_t, t)$, governing the probability flow ordinary differential equation (ODE):
\begin{equation}
\frac{d\boldsymbol{y}_t}{dt} = \boldsymbol{v}(\boldsymbol{y}_t, t).
\end{equation}
This velocity function captures the expected trajectory of $\boldsymbol{y}_t$ over time, derived as:
\begin{equation}
\boldsymbol{v}(\boldsymbol{y}_t, t) = \dot{a}_t \cdot \mathbb{E}[\boldsymbol{z} \mid \boldsymbol{y}_t] + \dot{b}_t \cdot \mathbb{E}[\boldsymbol{\epsilon} \mid \boldsymbol{y}_t],
\end{equation}
where $\dot{a}_t$ and $\dot{b}_t$ denote time derivatives of $a_t$ and $b_t$, respectively.

To learn $\boldsymbol{v}(\boldsymbol{y}_t, t)$, SiT trains a parameterized model $\boldsymbol{v}_{\boldsymbol{\phi}}(\boldsymbol{y}_t, t)$ to minimize the squared error between predicted and true velocities:
\begin{equation}
\mathcal{L}_{\boldsymbol{\phi}} =  \mathbb{E}_{t,\boldsymbol{z}, \boldsymbol{\epsilon}} \left[ \left\| \boldsymbol{v}_{\boldsymbol{\phi}}(\boldsymbol{y}_t, t) - \left( \dot{a}_t \boldsymbol{z} + \dot{b}_t \boldsymbol{\epsilon} \right) \right\|^2 \right] dt.
\end{equation}

For generation, one integrates the reverse of Eq. (2) starting from pure noise ($\boldsymbol{y}_T = \boldsymbol{\epsilon}$) to recover $\boldsymbol{z}$. Notably, SiT connects velocity to the score function $\boldsymbol{s}(\boldsymbol{y}_t, t)$ (used in stochastic differential equation (SDE) sampling) via:
\begin{equation}
\boldsymbol{s}(\boldsymbol{y}_t, t) = -\frac{1}{b_t} \mathbb{E}[\boldsymbol{\epsilon} \mid \boldsymbol{y}_t],
\end{equation}
which can be rewritten using the velocity function:
\begin{equation}
\boldsymbol{s}(\boldsymbol{y}_t, t) = \frac{\dot{a}_t \boldsymbol{y}_t - a_t \boldsymbol{v}(\boldsymbol{y}_t, t)}{{b_t}(a_t \dot{b_t} - \dot{a}_t b_t)}
\end{equation}

Vanilla SiT is trained with this denoising objective $\mathcal{L}_{\boldsymbol{\phi}}$, which enables learning a unified velocity function and provides a strong foundation for generative modeling as our baseline.  

\subsection{VAE Feature Alignment}
Vanilla SiT's velocity prediction relies solely on the denoising signal without leveraging visual priors guidance, a strategy that has been validated as highly effective in recent works~\cite{repa,sra,reg}. To identify a readily available visual feature source rich in informative cues but without introducing additional pre-trained feature encoders or online extra teacher diffusion models, we turn to the first-stage output of the LDM framework itself: the off-the-shelf features of the pre-trained VAE, whose reconstruction property ensures encoded texture details, structural patterns, and basic semantic information as shown in Fig.~\ref{fig:pca}. This motivates us to propose a novel VAE feature representation alignment framework to accelerate the training and enhance the generative fidelity of the vanilla diffusion models. As shown in Fig.~\ref{fig:paradigms}(d), our \name retains SiT’s core framework while introducing an alignment component to bridge VAE features and SiT’s latent space, requiring only the addition of a lightweight MLP.

\subsubsection{VAE Feature Extraction}
Since SiT is a latent generative model, it follows the two-stage LDM~\cite{ldm} training pipeline: the first stage trains a VAE to map raw images into a compact latent space, and the second stage trains a generative model (e.g., diffusion model) in this latent space. We follow SiT's setting and use the same pre-trained VAE from Stable Diffusion~\cite{sd-vae} (SD-VAE), leveraging its capability in learning meaningful, reconstruction-capable visual representations. 

For an input image $\boldsymbol{x}$, the SD-VAE encoder maps it to a compact latent embedding space as a VAE feature tensor $\boldsymbol{f}_{\text{VAE}} \in \mathbb{R}^{C \times H \times W}$ (with a shape of 4×32×32 for 3×256×256 input images), where $C$ denotes the feature channel dimension, and $H, W$ are the spatial dimensions of the feature maps. Notably, in the training process of the second-stage diffusion models in LDM, $\boldsymbol{f}_{\text{VAE}}$ is typically pre-extracted and stored for later use to eliminate the on-the-fly feature extraction during training~\cite{sit,dit,repa,reg}. Thus, we refer to these as off-the-shelf features.

\subsubsection{Latent Representation Alignment}
The core of our \name is to align intermediate features of SiT with the VAE’s prior-rich features during training. Let $\boldsymbol{y}_t$ denote SiT’s latent state at time step $t$, generated by the interpolating process $\boldsymbol{y}_t = a_t \boldsymbol{z} + b_t \boldsymbol{\epsilon}$. We extract an intermediate feature tensor $\boldsymbol{h}_{\text{SiT}}$ from a certain hidden layer of SiT, then project this intermediate feature into the same feature space as $\boldsymbol{f}_{\text{VAE}}$ using a lightweight MLP $\mathcal{P}(\cdot)$, yielding the aligned SiT feature $\boldsymbol{f}_{\text{SiT}} = \mathcal{P}(\boldsymbol{h}_{\text{SiT}})$, as shown in Fig.~\ref{fig:paradigms}(d).

To enforce alignment, we adopt a\textit{ smooth $\mathcal{L}_1$ loss} as the alignment objective $\mathcal{L}_{\text{align}}$ on the element-wised feature difference $ \boldsymbol{\Delta f} =\boldsymbol{f}_{\text{SiT}} - \boldsymbol{f}_{\text{VAE}}$ as:
\begin{equation}
\mathcal{L}_{\text{align}} = \mathbb{E}_{\boldsymbol{z}, \boldsymbol{\epsilon}, t} \left[ \sum_{i=1}^{N} 
\begin{cases} 
\frac{1}{2\beta} (\boldsymbol{\Delta f}_i)^2 & \text{if } | \boldsymbol{\Delta f}_i| \leq \beta, \\
\frac{|\boldsymbol{\Delta f}_i|}{\beta} - \frac{1}{2} & \text{otherwise}
\end{cases}
\right],
\end{equation}
where $N = C\times W \times H$ and $\beta$ controls the threshold between the quadratic and linear regions and is set to 0.05 in all our experiments, the sum is taken over all elements of $\boldsymbol{\Delta f}$, and the expectation is computed over the input $\boldsymbol{z}$, Gaussian noise $\boldsymbol{\epsilon}$, and time step $t$. This loss encourages SiT’s intermediate features to capture similar fine-grained details, structural patterns and semantic information as the VAE’s feature maps, infusing valuable visual priors into the diffusion learning process.

\subsubsection{Overall Training Objective}
The overall training objective of our \name framework is a weighted combination of vanilla SiT’s denoising loss and our proposed alignment loss:
\begin{equation}
\mathcal{L}_{\text{total}} = \mathcal{L}_{\boldsymbol{\phi}} + \lambda \cdot \mathcal{L}_{\text{align}},
\end{equation}
where $\lambda$ is a hyperparameter that balances the contribution of the two losses. By optimizing $\mathcal{L}_{\text{total}}$, the model retains SiT’s advantage while leveraging the VAE’s representations to refine latent space learning, ultimately accelerating the training process and improving the generation fidelity.
\section{Experiments}
\label{sec:experiment}

\begin{table*}[t]
\centering
\small
\setlength{\tabcolsep}{8pt}
\captionof{table}{
\textbf{Ablation studies} on ImageNet 256$\times$256 without classifier-free guidance (CFG), which employs SiT-B/2 architectures trained for 400K iterations (with a batch size of 256). $\downarrow$ and $\uparrow$ indicate whether lower or higher values are better, respectively.  
}
\resizebox{0.9\linewidth}{!}{
\begin{tabular}{c c c c c | c c c c c}
\toprule
Align Depth & Timesteps & Align Objective& $\lambda$ & MLP Setting  &  {\pz FID$\downarrow$} & {sFID$\downarrow$} & {IS$\uparrow$} & {Pre.$\uparrow$} & Rec.$\uparrow$ \\
\midrule
\rowcolor{gray!20}
\multicolumn{1}{l}{Vanilla SiT-B/2~\cite{sit}}&- &-&-&-& 33.02 &6.46 & 43.71 &0.53 &0.63\\
\midrule  		
\cellcolor{tabhighlight}2 & $[0, 1]$ &smooth-$\ell_1$ & 1.0 & 5-layer &\cellcolor{tabhighlight}28.89 &\cellcolor{tabhighlight}6.20 &\cellcolor{tabhighlight}51.64 &\cellcolor{tabhighlight}0.56	&\cellcolor{tabhighlight}0.63\\
\cellcolor{tabhighlight}3 & $[0, 1]$ &smooth-$\ell_1$& 1.0 & 5-layer & 29.11 & 6.07 &50.93 &0.56 &0.64 \\			
\cellcolor{tabhighlight}4 & $[0, 1]$ &smooth-$\ell_1$& 1.0 & 5-layer &29.51	&6.15	&50.50	&0.56	&0.65
 \\
\cellcolor{tabhighlight}6 & $[0, 1]$ &smooth-$\ell_1$& 1.0 & 5-layer & 32.44 &6.24	&45.65	&0.54	&0.64
 \\
\cellcolor{tabhighlight}8 & $[0, 1]$ &smooth-$\ell_1$& 1.0 & 5-layer &36.20 &6.50 & 40.88	&0.52	&0.64
 \\
\midrule
2 & \cellcolor{tabhighlight}$[0, 1]$ &smooth-$\ell_1$& 1.0& 5-layer &\cellcolor{tabhighlight}28.89 &\cellcolor{tabhighlight}6.20 &\cellcolor{tabhighlight}51.64 &\cellcolor{tabhighlight}0.56	&\cellcolor{tabhighlight}0.63 \\
2 & \cellcolor{tabhighlight}$[0, 0.5]$ &smooth-$\ell_1$& 1.0 & 5-layer &30.04 &6.24	&49.80	&0.56 &0.64
\\
2 & \cellcolor{tabhighlight}$[0.5, 1]$ &smooth-$\ell_1$& 1.0 & 5-layer &29.59	&6.18	&49.71	&0.56	&0.64
 \\
\midrule
2 & \pz$[0, 1]$ & \cellcolor{tabhighlight}smooth-$\ell_1$& 1.0& 5-layer &\cellcolor{tabhighlight}28.89 &\cellcolor{tabhighlight}6.20 &\cellcolor{tabhighlight}51.64 &\cellcolor{tabhighlight}0.56	&\cellcolor{tabhighlight}0.63 \\
2 & \pz$[0, 1]$ & \cellcolor{tabhighlight}cosine& 1.0 & 5-layer &29.30 &6.20&50.51	&0.56	&0.63
\\
2 & \pz$[0, 1]$ & \cellcolor{tabhighlight} $\ell_1$& 1.0 & 5-layer & 29.50 &6.20 &50.14 &0.56 &0.64  \\
2 & \pz$[0, 1]$ & \cellcolor{tabhighlight}$\ell_2$& 1.0 & 5-layer & 29.40 & 6.16 &50.46 &0.56 &0.64   \\
\midrule
2 & \pz$[0, 1]$  &smooth-$\ell_1$& \cellcolor{tabhighlight}1 & 5-layer &\cellcolor{tabhighlight}28.89 &\cellcolor{tabhighlight}6.20 &\cellcolor{tabhighlight}51.64 &\cellcolor{tabhighlight}0.56	&\cellcolor{tabhighlight}0.63\\
2 & \pz$[0, 1]$  &smooth-$\ell_1$& \cellcolor{tabhighlight}0.1 & 5-layer &30.10	&6.21	&50.38	&0.55	&0.64
 \\
2 & \pz$[0, 1]$  &smooth-$\ell_1$& \cellcolor{tabhighlight}0.5 & 5-layer &29.50	&6.19	&50.50	&0.56	&0.64
 \\
\midrule
2 & \pz$[0, 1]$  &smooth-$\ell_1$& 1.0 & \cellcolor{tabhighlight}5-layer &\cellcolor{tabhighlight}28.89 &\cellcolor{tabhighlight}6.20 &\cellcolor{tabhighlight}51.64 &\cellcolor{tabhighlight}0.56	&\cellcolor{tabhighlight}0.63\\
2 & \pz$[0, 1]$  &smooth-$\ell_1$& 1.0 & \cellcolor{tabhighlight}2-layer & 31.32  &6.51 &48.15 &0.55 &0.64  \\
\bottomrule
\end{tabular}
}
\label{tab:ab}
\end{table*}
\subsection{Experimental Setup}
\textbf{Implementation details.} All experiments are performed on the ImageNet dataset~\cite{imagenet}, where images are preprocessed to 256×256 resolution through center cropping and resizing. To ensure comparability, our training settings strictly follow the configurations specified in SiT~\cite{sit} and REPA~\cite{repa}. This encompasses the AdamW optimizer~\cite{adamw} with a constant learning rate of 1e-4, no weight decay, a fixed batch size of 256, and the use of SD-VAE~\cite{sd-vae} for latent VAE feature extraction. Regarding model architectures, we employ the B/2, L/2, and XL/2 designs from SiT, which process inputs with a patch size of 2. Sampling follows the SDE Euler–Maruyama solver with 250 steps, as in REPA~\cite{repa}. Additional implementation specifics are available in the Appendix.

\noindent\textbf{Evaluation protocol.} To comprehensively assess image generation quality across multiple dimensions, we employ a rigorous set of quantitative metrics, including Fréchet Inception Distance (FID)~\cite{fid} for realism, structural FID (sFID)~\cite{sfid} for spatial coherence, Inception Score (IS)~\cite{is} for class-conditional diversity, precision (Pre.) and recall (Rec.)~\cite{pre-rec} for sample fidelity and target distribution coverage, respectively. All metrics are computed on a standardized set of 50K generated samples to ensure statistical reliability~\cite{vavae,sra,reg,repa}. 

\subsection{Ablation Studies and Discussions}
\noindent\textbf{Alignment Depth.} We investigate the effects of applying our \name at different network depths in Tab.~\ref{tab:ab}. Our analysis reveals that applying the \name alignment in earlier layers yields superior results, which is also consistent with previous works'~\cite{repa,sra,reg} findings. Performance gradually degrades as the alignment shifts to deeper layers; we hypothesize this is because deeper layers focus on precise fine-grained details or semantic information, which go beyond the ability of VAE features, and imposing constraints here disrupts their natural refinement of such details.
Notably, our method demonstrates the best results in layer 2, achieving an FID of 28.89, a reduction of 4.13 compared to the baseline. Based on this observation, we strategically apply \name alignment at layers 2, 8, and 8 for the B, L, and XL architectures, respectively.

\noindent\textbf{Timesteps.} We also examine \namex's effect across timestep ranges in Tab.~\ref{tab:ab}. Our analysis shows applying alignment over the full range \( t \in [0, 1] \) yields the best performance, outperforming both \( t \in [0, 0.5] \) and \( t \in [0.5, 1] \).  
This is due to complementary alignment across noise levels: in low-noise stages, \name refines coherent representations using the VAE's texture details and structural patterns; in high-noise stages, it helps the model against degradation via the VAE's visual properties. Covering \( [0, 1] \) ensures the VAE's rich features are leveraged consistently throughout the entire diffusion process. Thus, we adopt the full timestep range \( t \in [0, 1] \) for \name to fully utilize the guidance of VAE features.

\noindent\textbf{Alignment Objective.} Next, we evaluate the impact of different alignment objectives in Tab.~\ref{tab:ab}, including \( \ell_1 \), \( \ell_2 \), and \( \text{smooth-}\ell_1 \). The results show that the \( \text{smooth-}\ell_1 \) loss achieves the best overall performance, while the others also perform reasonably well. Consequently, this objective was adopted as the default for all subsequent experiments.

\noindent\textbf{Effect of $\lambda$.} We further explore the impact of the alignment loss weight \( \lambda \) in Tab.~\ref{tab:ab}. Among the tested values, \( \lambda = 1.0 \) achieves the best overall performance across the major evaluation metrics. Therefore, we set \( \lambda = 1.0 \) as the default parameter in our experiments.

\noindent\textbf{MLP Setting.} We finally demonstrate the impact of different MLP layers in Tab.~\ref{tab:ab}. The results show that using a 2-layer MLP (1M) yields suboptimal performance, while a 5-layer MLP (8M) achieves promising results. We believe that there exists a significant feature space discrepancy between SiT features and VAE features, as illustrated in Fig.~\ref{fig:pca}, and a deeper MLP can better perform the feature transformations needed to align with the VAE's features, thus enabling more effective feature integration and refinement throughout the diffusion process. We apply a 5-layer MLP as the default setting for subsequent experiments.
\begin{figure*}
  \centering
  \includegraphics[width=1\linewidth]{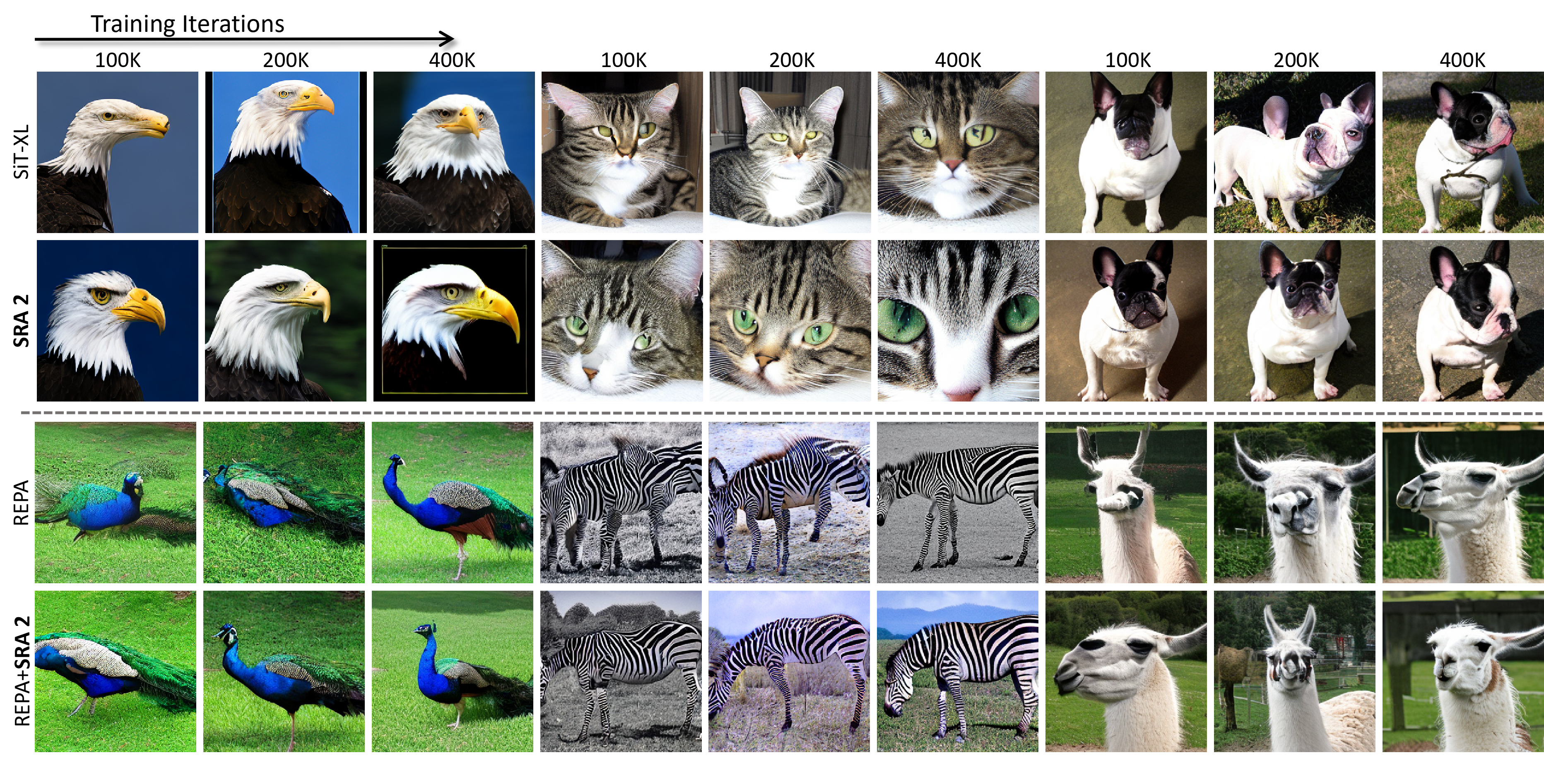} 
  \vspace{-.7cm}
\caption{\textbf{SRA 2 Improves Visual Scaling.} \textbf{Top}: Vanilla SiT-XL/2 and SiT-XL/2+\namex. \textbf{Bottom}: Vanilla REPA and REPA+\namex. Our method is verified to produce images with higher structural fidelity, finer details, and stronger semantic coherence at the same training steps compared with both vanilla SiT and vanilla REPA. Results for all methods are sampled using the same seed, noise, and class label, with a classifier-free guidance scale of 4.0 employed during sampling.}\label{fig:sample_compare}
  \vspace{-.4cm}
\end{figure*}
\subsection{System-level Comparisons}
\noindent\textbf{Accelerating training convergence.} Tab.~\ref{tab:wo_cfg} provides a detailed comparison of training convergence across different model scales and iterations on ImageNet 256×256 without classifier-free guidance (CFG). The proposed \name method consistently accelerates training convergence while reducing FID scores. For SiT-B/2, our \name achieves an FID of 28.9 at 400K iterations, outperforming the baseline SiT-B/2 by 4.1 points. For SiT-L/2, \name reaches an FID of 14.3 at 400K iterations, surpassing SiT-L/2’s 18.8 at 400K and even SiT-XL/2’s 14.6 at 600K. For SiT-XL/2, \name gets an FID of 8.2 at 1M iterations, better than SiT-XL/2’s 8.3 at 7M, representing a 7$\times$ training acceleration for better performance. Moreover, \name continues to improve, reaching 6.6 at 4M iterations. This demonstrates that our \name effectively accelerates training convergence across different model scales. Additionally, we present the generation quality comparison at different training steps in Fig.~\ref{fig:sample_compare}, where \name stably enhances generation quality, further validating its effectiveness.

\noindent\textbf{Compatibility with Other Methods.} Tab.~\ref{tab:wo_cfg} also shows the compatibility of \name with other methods. When combined with SiT-XL/2+REPA~\cite{repa}, using \name achieves consistent improvements across 100K, 200K, and 400K iterations, with FID score reductions of 3.1, 1.9, and 1.1, respectively, outperforming SiT-XL/2+REPA at each corresponding iteration. Additionally, when integrated with VAVAE~\cite{vavae}, \name further reduces the FID to 4.4 at 400K iterations from 4.9, indicating that integrating our method with other advanced methods is compatible and can bring additional performance gains. Qualitatively, we present the generation quality of vanilla REPA and REPA+\name at 100K, 200K, and 400K iterations in Fig.~\ref{fig:sample_compare}. On top of REPA, \name generates more realistic images with more plausible details, further validating its compatibility.
\begin{figure*}[t]
\vspace{0.05in}
\begin{minipage}{0.4\textwidth}
\centering\small
\captionof{table}{\textbf{FID comparison across training iterations for accelerated alignment methods.} All experiments are conducted on ImageNet (256×256) with a batch size of 256 and without CFG. }
\resizebox{0.95\textwidth}{!}{%
\begin{tabular}{lccc}
\toprule
     Method  & \#Params & Iter. & FID$\downarrow$  \\
     \midrule
     SiT-B/2  &130M & 400K & 33.0 \\
     {\textbf{+ \namex (ours)}} &130M & \textbf{400K}  & \textbf{28.9}\\
     \arrayrulecolor{black!40}\midrule
     SiT-L/2 &458M & 200K & 28.2 \\
     SiT-L/2 &458M & 400K & 18.8 \\
     {\textbf{+ \namex (ours)}} &458M & \textbf{200K}  & \textbf{22.5}\\
     {\textbf{+ \namex (ours)}} &458M & \textbf{400K} & \textbf{14.3} \\
     \arrayrulecolor{black!40}\midrule
     SiT-XL/2  &675M & 400K   & 17.2\\
     SiT-XL/2  &675M & 600K   & 14.6\\
     SiT-XL/2  &675M & 7M   & 8.3\\
     {\textbf{+ \namex (ours)}} &675M & \textbf{400K}  & \textbf{11.7}\\
     {\textbf{+ \namex (ours)}} &675M & \textbf{1M}  & \textbf{8.2}\\
     {\textbf{+ \namex (ours)}} &675M & \textbf{2M}  & \textbf{7.0}\\
     {\textbf{+ \namex (ours)}} &675M & \textbf{4M}  & \textbf{6.6}\\
     \arrayrulecolor{black}\midrule
     {SiT-XL/2 + REPA~\cite{repa}} &675M & 100K  & 19.4\\
     {SiT-XL/2 + REPA~\cite{repa}} &675M & 200K  & 11.1\\
     {SiT-XL/2 + REPA~\cite{repa}} &675M & 400K  & 7.9\\
     {\textbf{+ \namex (ours)}} &675M & \textbf{100K}  & \textbf{16.3}\\
     {\textbf{+ \namex (ours)}} &675M & \textbf{200K}  & \textbf{9.2}\\
     {\textbf{+ \namex (ours)}} &675M & \textbf{400K}  & \textbf{6.8}\\\midrule
     {VAVAE~\cite{vavae}} &675M & 400K  & 4.9 \\
     {\textbf{+ \namex (ours)}} & 675M & \textbf{400K}  & \textbf{4.4}\\
     \bottomrule
\end{tabular}
\label{tab:wo_cfg}
}
\end{minipage}
~~
\begin{minipage}{0.58\textwidth}

\centering\large
\captionof{table}{\textbf{Comparison of the performance of different methods on ImageNet 256$\times$256 with CFG.} Performance metrics are annotated with $\uparrow $ (higher is better) and $\downarrow$ (lower is better). ``External" indicates whether they rely on external dependencies (e.g., extra encoders, diffusion teachers, or diffusion decoders). }
\vspace{-0.05in}
\resizebox{1\textwidth}{!}{%
\begin{tabular}{l c c c c c c c}
\toprule
{ Method} & Epochs  &  { FID$\downarrow$} & {sFID$\downarrow$} & {IS$\uparrow$} & {Pre.$\uparrow$} & Rec.$\uparrow$ & External \\
\arrayrulecolor{black}\midrule

\multicolumn{8}{l}{\emph{Pixel diffusion}\vspace{0.02in}} \\
 {ADM-U~\cite{adm}}  &400 &  3.94 & 6.14 &  186.7 & 0.82 & 0.52 &{\color{red}\ding{55}} \\
 VDM$++$~\cite{vdm} & 560 & 2.40 & - &  225.3 & - & - &{\color{red}\ding{55}}  \\
 Simple diffusion~\cite{simplediff} & 800 & 2.77 & - & 211.8 & - & - &{\color{red}\ding{55}} \\
 CDM~\cite{cdm}  & 2160 & 4.88 & - & 158.7 & - & - \\
\arrayrulecolor{black!40}\midrule

\multicolumn{8}{l}{\emph{Latent diffusion, U-Net}\vspace{0.02in}} \\
 LDM-4~\cite{ldm} & 200  & 3.60 & - & 247.7 & {0.87} & 0.48 &{\color{red}\ding{55}}\\
\arrayrulecolor{black!40}\midrule

\multicolumn{8}{l}{\emph{Latent diffusion, Transformer + U-Net hybrid}\vspace{0.02in}} \\
 U-ViT-H/2~\cite{uvit} & 240 & 2.29 & 5.68  & 263.9 & 0.82 & 0.57&{\color{red}\ding{55}} \\ 
 DiffiT~\cite{diffit} & - & 1.73 & - &  276.5 & 0.80 & 0.62 &{\color{red}\ding{55}} \\
\arrayrulecolor{black} \midrule

\multicolumn{7}{l}{\emph{Latent diffusion, Transformer}\vspace{0.02in}} \\
 {DiT-XL/2}~\cite{dit}   & 1400  &    2.27 & 4.60 & {278.2} & {\textbf{0.83}} & 0.57 &{\color{red}\ding{55}} \\
{{{+ iSSD}~\cite{ssd}}} & 1400+0.8 &{2.02} & {4.22} & {250.0} & {0.81} & {{0.60}} &{\color{green}\ding{51}}\\
 MaskDiT~\cite{maskdit} & 1600 &  2.28 & 5.67 & 276.6 & 0.80 & 0.61 &{\color{green}\ding{51}} \\ 
 SD-DiT~\cite{sddit} & 480 & 3.23 & -    & -     & -    & -   &{\color{green}\ding{51}}   \\
\arrayrulecolor{black!30}\cmidrule(lr){1-8}
 {SiT-XL/2}~\cite{sit}    & 1400 &     2.06 & {4.50} & 270.3 & 0.82 & 0.59 &{\color{red}\ding{55}} \\
{{{+ SRA}~\cite{sra}}} & {{800}} &{1.58} & {4.65} & {{305.7}} & {0.80} & {{0.65}} &{\color{green}\ding{51}}\\
{{{+ REPA}~\cite{repa}}} & {{800}} &{1.42} & {4.70} & {{311.4}} & {0.80} & {{0.63}} &{\color{green}\ding{51}} \\
{{+ REG}~\cite{reg}} & {800} &\textbf{1.36}  &4.25  &299.4  &0.77  &\textbf{0.66} &{\color{green}\ding{51}}\\
{{\textbf{+ \namex (ours)}}} & {100} & {2.39} & {{4.54}} & {252.5} & \textbf{0.83} & {0.56} &{\color{red}\ding{55}} \\
{{\textbf{+ \namex (ours)}}} & {200} & {1.98} &{{4.53}} & {{284.5}} & \textbf{0.83} & {0.59}&{\color{red}\ding{55}} \\
{{\textbf{+ \namex (ours)}}} & {400}   &1.84  &4.50  &306.5  &0.82 &0.61 &{\color{red}\ding{55}}\\
{{\textbf{+ \namex (ours)}}} & {800} &1.52  &4.63  &\textbf{316.2}  &0.82  &0.62 &{\color{red}\ding{55}} \\
\arrayrulecolor{black}\bottomrule
\end{tabular}
}
\label{tab:main}
\end{minipage}
\vspace{-2mm}
\end{figure*}

\noindent\textbf{Comparison with SOTA methods.} Tab.~\ref{tab:main} presents a comprehensive comparison against recent SOTA methods on ImageNet 256×256 with CFG, focusing on differences in external dependencies. Our proposed \name method achieves competitive performance while demonstrating efficient training convergence \textbf{without any external dependencies}. Against methods with external diffusion decoders (e.g., MaskDiT~\cite{maskdit}, SD-DiT~\cite{sddit}), \name surpasses MaskDiT’s performance at 1300 epochs with an FID of 1.98 at only 200 epochs, and outperforms SD-DiT’s result at 480 epochs using just 100 epochs. Against methods with external pretrained encoders (e.g., REPA~\cite{repa}, REG~\cite{reg}), \name achieves a comparable FID (1.52 vs. REPA’s 1.42 at 800 epochs) and superior IS (316.2 vs. REPA’s 311.4). Against methods with external teacher diffusion models (e.g., SRA~\cite{sra}), \name maintains better performance on both FID and IS at the same epochs. Additionally, \name shows consistent improvement with increasing training epochs, validating its robustness and scalability. These results highlight that our method can compete with SOTA approaches across different external dependency paradigms while maintaining training efficiency and architectural simplicity. We provide selected qualitative results of SiT-XL/2+\namex in Fig.~\ref{fig:samples} and additional examples in the Appendix, all of which exhibit excellent image quality.
\begin{table}[htbp]
\centering \small
\caption{\textbf{Generalization to T2I Tasks.} We find that SRA 2 also generalizes to T2I tasks, yielding improved generation performance.}
\label{tab:t2i_results}
\begin{tabular}{l|c|c}
\hline
Method & FID $ \downarrow $  & PickScore $ \uparrow $  \\
\hline
MMDiT Baseline~\cite{mmdit} & 5.08 & 20.54 \\
\hline
MMDiT + REPA~\cite{repa} & 4.60 & 20.90 \\
\hline
MMDiT + SRA~\cite{sra} & 4.74 & 21.12 \\
\hline
MMDiT + SRA 2 & 4.67 & 20.92 \\
\hline
\end{tabular}
\end{table}

\noindent\textbf{Text-to-Image Generation Experiment.}
We validate SRA 2 in text-to-image (T2I) generation on MS-COCO~\cite{mscoco}, following the experimental protocol of REPA~\cite{repa}. Specifically, we adopt MMDiT~\cite{mmdit} as the diffusion backbone and train all models for 150K iterations with a batch size of 256. During inference, we apply classifier-free guidance with a scale of 2.0. As shown in Tab.~\ref{tab:t2i_results}, SRA 2 achieves competitive performance, yielding a FID of 4.67 and a PickScore of 20.92, surpassing the baseline and matching or approaching methods that rely on external representation learners, despite using only the built-in VAE. This demonstrates that our method can effectively generalize to T2I generation tasks.
\begin{table}[t]
    \centering
     \caption{\textbf{Training computational computational cost comparison.} This table compares REPA, SRA, and \name on ImageNet 256$\times$256, detailing  external forward parameters (EFP, formatted as \textcolor{Rhodamine}{external model parameters} + \textcolor{cyan}{MLP head parameters}), training speed per batch (size 256) (TS), GFLOPs, and forward latency. Values in \textcolor{red}{red} parentheses indicate changes relative to the SiT-XL/2 baseline. These results were tested on H100 GPUs.}
    \scriptsize
    \resizebox{1\linewidth}{!}
    {
      \begin{tabular}{l|cccc}
        \hline
        Method &\#EFP(M) &TS (Iter/s)$\uparrow$ & GFLOPs$\downarrow$  &Latency (ms)$\downarrow$  \\
        \hline
        SiT-XL/2 &\textcolor{Rhodamine}{0~}+\textcolor{cyan}{~0} &8.1 & 114.46   & 18.05 \\
        +REPA~\cite{repa} & \textcolor{Rhodamine}{86~}+\textcolor{cyan}{~8} & 6.3 (\textcolor{red}{-22\%})  & 138.50 (\textcolor{red}{+21\%})   & 22.81 (\textcolor{red}{+26\%})  \\
        +SRA~\cite{sra}  &\textcolor{Rhodamine}{481~}+\textcolor{cyan}{~2} &5.1(\textcolor{red}{-37\%}) & 197.57 (\textcolor{red}{+73\%})   & 30.90 (\textcolor{red}{+71\%}) \\
        \hline
        +\name  &\textcolor{Rhodamine}{0~}+\textcolor{cyan}{~18} &7.2(\textcolor{red}{\textbf{-11\%}}) & 118.55 (\textcolor{red}{\textbf{+4\%}})  & 19.20 (\textcolor{red}{\textbf{+6\%}})  \\
        \hline
      \end{tabular}
    }
    \label{tab:flops}
    \vspace{-5mm}
\end{table}

\noindent\textbf{Training Computational Cost Comparison.} Tab.~\ref{tab:flops} compares the training computational cost of typical extra-encoder-required REPA~\cite{repa}, dual-model SRA~\cite{sra}, and our \name on ImageNet 256×256. Among the three methods, our \name demonstrates the most significant advantages in computational efficiency: it requires 0 external model parameters, in stark contrast to REPA (86M) and SRA (481M). In terms of training speed, \name maintains performance close to the baseline SiT-XL/2 (only 11\% slower per batch), outperforming REPA (-22\%) and SRA (-37\%), which suffer larger slowdowns. For GFLOPs, \name adds merely 4\% compared to the baseline, far superior to the substantial increments of REPA (+21\%) and SRA (+73\%). Regarding forward latency, \name introduces only a 6\% increase relative to the baseline (attributed to the lightweight MLP alignment head), while REPA and SRA show much larger overheads (+26\% and +71\%, respectively). These results highlight \namex’s remarkable training efficiency, with no external encoder parameters and minimal increments in training speed, GFLOPs and latency, making it significantly more lightweight than existing methods.

\section{Conclusion}
\label{sec:conclusion}
This work addresses the training acceleration  of diffusion transformers. We propose \namex, a lightweight framework that leverages off-the-shelf pre-trained SD-VAE features for representation alignment. These features are inherently rich in texture details, structural patterns, and basic semantic information. Unlike existing methods that rely on external encoders that are not available in all domains or dual-model setups, \name reuses pre-extracted VAE features and only introduces a lightweight projection layer and an alignment loss. 
Experiments confirm that \name accelerates training, enhances generation quality, and complements other methods, all with minimal additional computational overhead.
This work demonstrates that pre-trained VAE visual priors are a powerful low-cost resource for efficient diffusion training. It offers a practical path to balancing efficiency and generation quality.

\section*{Acknowledgment}
This work was supported by the National Natural Science Foundation of China under Grant No.~62403429, No.~62476247, the Hangzhou Key Research and Development Program under Grant 2025SZDA0100, and the Zhejiang Provincial Natural Science Foundation of China under Grant No.~LQN25F030008.
{
    \small
    \bibliographystyle{ieeenat_fullname}
    \bibliography{main}
}

\end{document}